\documentclass[runningheads,a4paper]{llncs}
\usepackage{amssymb}
\usepackage{graphicx}
\usepackage{pgfplots}
\usepackage{algorithm2e}
\usepackage{mathtools,breqn}
\usepackage{rotating}
\usepackage{multirow}
\usepackage{comment}

\usepackage{url}
\urldef{\mailsa}\path|{alfred.hofmann, ursula.barth, ingrid.haas, frank.holzwarth,|
\urldef{\mailsb}\path|anna.kramer, leonie.kunz, christine.reiss, nicole.sator,|
\urldef{\mailsc}\path|erika.siebert-cole, peter.strasser, lncs}@springer.com|

\newcommand{\plotffind}[5]{
	\begin{axis}[
		minor tick num=3,
		axis y line=center,
		axis x line=middle,
		enlargelimits = true,
		xmax =7,
		no markers,
		ylabel={#4}, xlabel={#2},   
		y tick label style={/pgf/number format/.cd,%
			scaled y ticks = false,
			set thousands separator={},
			fixed},
		x tick label style={/pgf/number format/.cd,%
			scaled x ticks = false,
			set decimal separator={,},
			fixed}
		]   
		\addplot[very thick,blue,smooth] table[x index=#1,y index=#3] {#5};
	\end{axis}   
}

\tikzstyle{summe}=[circle,draw=blue!50,fill=blue!20,thick]
\tikzset{add/.style n args={4}{
    minimum width=6mm,
    path picture={
        \draw[black] 
            (path picture bounding box.south east) -- (path picture bounding box.north west)
            (path picture bounding box.south west) -- (path picture bounding box.north east);
        \node at ($(path picture bounding box.south)+(0,0.13)$)     {\tiny #1};
        \node at ($(path picture bounding box.west)+(0.13,0)$)      {\tiny #2};
        \node at ($(path picture bounding box.north)+(0,-0.13)$)        {\tiny #3};
        \node at ($(path picture bounding box.east)+(-0.13,0)$)     {\tiny #4};
        }
    }
}

\tikzstyle{object}=[rectangle,draw=black!50,fill=black!20,thick]

\begin{document}
\title{Neural network feedback controller for inertial platform}

\author{Yan Anisimov, Alexandr Lysov, Dmitry Kacai}

\institute{South Ural State University,\\
Lenina st. 86,  Chelyabinsk, Russia\\
\url{yan@yanchick.org}\\
\url{http://susu.ac.ru}}

\maketitle

\begin{abstract}
The paper describes an algorithm for the synthesis of neural networks to control gyro stabilizer. The neural network performs the role of observer for state vector. The role of an observer in a feedback of gyro stabilizer is illustrated. Paper detail a problem specific features stage of classics algorithm: choosing of network architecture, learning of neural network and verification of result feedback control. In the article presented optimal configuration of the neural network like a memory depth, the number of layers and neuron in these layers and activation functions in layers.  Using the information of dynamic system for improvement learning of neural network is provided.  A scheme creation of an optimal training sample is provided.
\end{abstract}

\section{Introduction}

Multilayer neural network can approximate any smooth function. For example, the transient of a state variables gyro stabilizer or control by time. The main advantage of a neural network is not requiring of a complete mathematical model of  gyro stabilizer. An example of these case is using MEMS gyroscopes for creating gyro stabilizer.  But, general algorithm of the synthesis of neural network not formulated yet. Such of algorithms is the goal for scientists. This paper is describing a development of this algorithm for the class of dynamic systems.

\section{Problem definition}

Dynamic a channel of uniaxial gyro stabilizer can be described by the following system of nonlinear differential equations \cite{IEEEhowto:Lysov}:

\begin{dmath}
\begin{array}{c}
A_1\ddot{\alpha_1} -\frac{(J_{xp} -J_{yp})} {2}\ddot{{\alpha }}_2 \cos \alpha_{2} \sin 2\alpha_{2}+\\+h\dot{{\alpha }}_1 + (J_{ze} -J_{xp})\dot{\alpha_1}\,\dot{\alpha_2}   =M_{1} +M_{\hbox{cont.}1}; \\

A_2 \ddot{{\alpha}}_2-\frac{(J_{\hbox{ye}}  -J_{\hbox{xe}} )} {2} \ddot{{\alpha }}_1 \cos \alpha_{2} \sin 2\alpha 
_{2} + \\+h \dot{{\alpha }}_2 + (J_{xp} -J_{ye})\dot{\alpha_3}\,\dot{\alpha_1}  =M_{2} + M_{\hbox{cont.}2} ; \\

A_3 \ddot{{\alpha }}_3  -J_{\hbox{ze}}
\ddot{{\alpha }}_1 \sin \alpha_{3} +\\+ h_3 \dot{{\alpha }}_3 +  (J_{zp} -J_{yi})\dot{\alpha_2}\,\dot{\alpha_3}=M_{3} +M_{\hbox{cont.}3}  ;\\

A_1 =J_{\hbox{ye}} +J_{\hbox{уi}} \cos^2\alpha_{2} +J_{\hbox{zi}} \sin ^2\alpha_{2} + \\ + J_{\hbox{yн}} \cos^2\alpha_{2}\cos^2\alpha_{3} +J_{\hbox{xp}} 
\cos^2\alpha_{2} \sin^2\alpha_{3} ;\\

A_2 =J_{\hbox{xi}} +J_{\hbox{xp}} \cos^2\alpha_{3} +J_{\hbox{yp}} \sin^2\alpha 
_{3} ;\\

A_3 = J_{\hbox{zp}};\\
\end{array}
\label{eq:platf}
\end{dmath}

where  $H$- Kinetic moment gyro unit, $J_{xp},J_{yp},J_{zp}$ - inertia moment of  platform,   $J_{xi},J_{yi},J_{zi}$ - inertia moment of  internal frame,   $J_{xe},J_{ye},J_{ze}$ - inertia moment of external frame, $h$ - damping factor,  $\alpha_1,\alpha_2,\alpha_3$- angle pumping platform, internal frame and external frame.

The nonlinear system \ref{eq:platf} append equation of inertial measurement unit:

\begin{equation}
\begin{array}{c}
\textbf{u}_g = \textbf{f}_g(\dot{\alpha}_1,\dot{\alpha}_2,\dot{\alpha}_3),  \\
\textbf{u}_a = \textbf{f}_a(\gamma_1,\gamma_2,\gamma_3),
\end{array}
\label{eq:imu_g}
\end{equation}

where $u_g$ - the signal from gyroscope, $u_a$ - signal of accelerometr. $\gamma_i$ -  angle between platform and horizone plane.  The model of IMU \ref{eq:imu_g} have a some features:

\begin{itemize}
\item the model have a dynamic properties;
\item the model is nonlinear;
\item the model is incomplete.

\end{itemize}

The problem of generating a control torque for compensation of the external torque by mesuared of signal of gyro($\textbf{u}_g$). Control is expected to form the law gyro stabilizer $M_{con.}=g(\alpha_1,\alpha_2,\alpha_3)$.  Where $M_{con.}$ is nonlinear dynamic link that is in a feedback loop. There some types of feedback controllers:

\begin{itemize}
	\item correcting unit;
	\item observer with regulator;
	\item neural network.
\end{itemize}

We consider case, when the feedback loop contains a neural network.

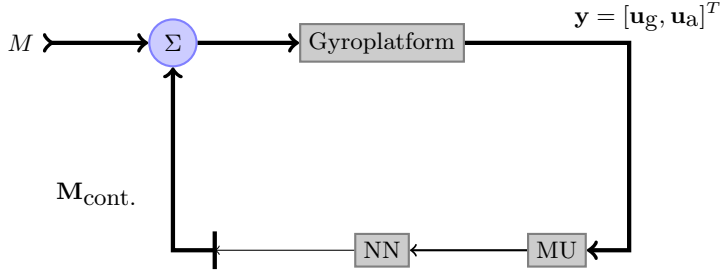
\begin{figure}
	\begin{tikzpicture}[matrix=true]
	\newcommand{\Gitter}[4]{
	    \draw[very thin,color=gray] (#1,#3) grid (#2,#4);
	    \node at (#1,#3)    {(#1,#3)};
	    \node at (#2,#4)    {(#2,#4)};
	    \node at (0,0)      {(0,0)};
	
	}
	
	\node at (2,7) (xix)     {$M$};
	\node at (4,7) (Sum1x) [summe] {$\Sigma$};
	\node at (6.75,7) (Intx)  [object] {{Gyroplatform}};
	\draw [>->]  [ultra thick]  (xix) -- (Sum1x.west);
	\draw [->]  [ultra thick]  (Sum1x.east)  -- (Intx.west);
	

	\node at (10.25,7.35) (yxx)     {\begin{small}$\textbf{y}= [\textbf{u}_{\hbox{g}},\textbf{u}_{\hbox{a}}]^T$\end{small}};
	\node at (9.05,4.25) (Bp)    [object] {{MU}};

	\node at (6.75,4.25) (K2)    [object] {{NN}};
	\node at (3.0,5) (yxx)   {{$\textbf{M}_{\hbox{cont.}}$}};
	
	\draw[ultra thick] [->] (Intx.east) -- (10,7) |- (Bp);
	\draw [->] [thick] (Bp) -- (K2);
	\draw [ultra thick]  (4.55,4.0) -- (4.55,4.5);
	\draw [->] (K2)-- (4.55,4.25);

	\draw [->] [ultra thick] (4.55,4.25) -- (4,4.25) -- (Sum1x);
	\end{tikzpicture}
	
	\caption{The structure of the control system. Neural network as regulator.}
\end{figure}

\begin{figure}
	\begin{tikzpicture}[matrix=true]

	\node at (2,7) (xix)     {$M$};
	\node at (4,7) (Sum1x) [summe] {$\Sigma$};
	\node at (6.75,7) (Intx)  [object] {{Gyroplatform}};
	\draw [>->]  [ultra thick]  (xix) -- (Sum1x.west);
	\draw [->]  [ultra thick]  (Sum1x.east)  -- (Intx.west);
	\node at (10.25,7.35) (yxx)     {\begin{small}$\textbf{y}= [\textbf{u}_{\hbox{g}},\textbf{u}_{\hbox{a}}]^T$\end{small}};
	\node at (6.55,4.65) (Ip)     {$\hat{\textbf{x}(t)}$};
	\node at (9.25,4.25) (Bp)    [object] {{ MU}};
	\node at (5.55,4.25) (Reg)    [object] {{ P}};
	
	\node at (7.75,4.25) (K2)    [object] {{NN}};
	\node at (3.0,5) (yxx)   {{$\textbf{M}_{\hbox{cont.}}$}};
	
	\draw[ultra thick] [->] (Intx.east) -- (10,7) |- (Bp);
	\draw [->] [thick] (Bp) -- (K2);
	\draw [->]  [thick] (K2)-- (Reg);
	\draw [->] [ultra thick] (Reg) -- (4,4.25) -- (Sum1x);
	\end{tikzpicture}
	\caption{The structure of the control system. Neural network as observer.}
\end{figure}
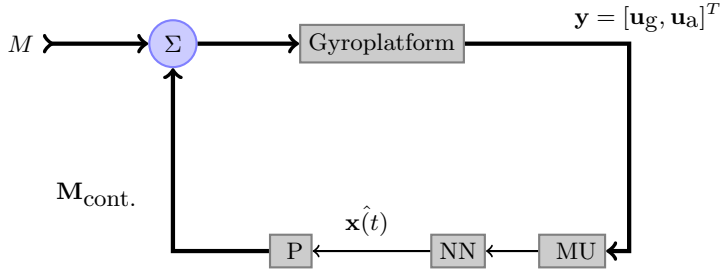

Two schemes of control are proposed. In the first scheme gyro signal put into the memory unit. The memory unit is generated a vector containing the current value and the previous several values of the vector $\hat{x}$. This vector $\hat{x}$ is input for neural network \cite{IEEEhowto:yan}.  

The neural network connected to a motor. The motor creates a moment, which balance external moment (Figure 1). In the second case, neural network estimate of the state vector, which is connected to the regulator, the signal from regulator is connected to the motors of stabilization (Figure 2).

In the case when gyro stabilizer has a several channels of stabilization feedback loop consists of several of parallel neural networks. This allow reduce the load on each neural network.

\section{Optimizing algorithm of synthesis}

We consider the "classic" algorithm for the synthesis of neural networks, which consists of five stages. In the first stage, formalization of the problem. The unknown function is determined that the neural network during its work will be interpolated, the number of input and output variables. Next the step is selecting structure of the neural network: definition of topology and network settings, types activation functions. After, creation of the training sample is following, which should reflect all the possible modes. Next step is  a choice of algorithm training parameters and train neural network. And the final stage is verification of the trained neural network on the test sample. When a result of checking is positive neural network is considered trained and may be used in the work.

Describe the algorithm for the synthesis of the control device consists of a neural network and regulator. So that the system (1) will be defined as:

\begin{equation}
u=-Px,
\end{equation}

where  $x$- State vector, $P$ - regulator.

\subsection{The formalization of the problem}

We proposed formulation of the problem a neural network works as an observer. The input of the neural network is vector of the measured signal and several previous values of it, and the output of the neural network - estimation of the state vector.

For base topology is selected multilayer neural network ("multilayer perceptron"). Mathematical model of the network is described by the equation \cite{IEEEhowto:Ossowski}:

\begin{eqnarray}
\begin{array}{l}
\widehat{\textbf{x}}(u_{\hbox{g.}})=f^{(n)} ...(f^{(2)}(\textbf{w}^{(2)}(f^{(1)}(^{(1)} [u_{\hbox{g.}}(k)\,u_{\hbox{g.}} (k-1)\,\cdots \\
\phantom{aa}\cdots\,u_{\hbox{g.}}(k-m) ]  ^{T} 
+\textbf{b}^{(1)} )+\textbf{b}^{(2)})...\textbf{b}^{(n)}_{i,0}),
\end{array}
\label{eq:MnogosloyDynamic}
\end{eqnarray}

where $w^{(j)}$  - Weighting matrix $j$-th layer of the neural network,  $b^{(j)}$- bias vector of $j$-th layer of the neural network,  $f^{(j)}$ - activation function $j$-th layer of the neural network, $u_{\hbox{g.}}(k)$  -current value of the measured signal,$\widehat{\textbf{x}}$  - An output vector of the neural network,  $k$- depth of memory.

Tables \ref{tab:max_angel_logsig},\ref{tab:max_angel_sig} are provided the result of simulation gyro stabilizer. Many of neural networks were synthesized during experiments for detecting the relationship between the parameters of the neural network and the features of the transient process in the stabilization of the platform. The tables show not all network can work as observer. Sometimes the transient process is unstable($\inf$ in tables).  Stable transient process can be find when correct inequality: 
  
 \begin{equation}
 m<k
 \label{eq:ineq}
 \end{equation}

 where  $m$- number neutral in hidden layer.
 
 The sub optimal, in case minimization of angle plumping, estimate when numbers neural in hidden layer is approximate equal to order of system (1)-(2). So, inequality \ref{eq:ineq} can be appended:
 
  \begin{equation}
  m \approx order(System)
  \label{eq:ineq}
  \end{equation}

 
 Neural network with nonlinear activation, like “tansig” or “logsig” function in hidden layer and linear in output layer are preferred. These features also work when the neural network is used for observing a linear dynamic system.
 
 \subsection{Creating a training sample}
 
 The training set is prepared with a special algorithm. The main aim is all system state variables are observable. Next, a closed system is formed by including a feedback loop controller by state. In some works are recommended use a harmonic signal with increasing frequency to the input of gyroscope stabilizer. But most prefer is use random normalized input signal or a harmonic signal at a fixed frequency. These results were obtained on the basis of numerical modeling.

 \begin{figure}[h!]
 	\begin{center}
 		\begin{tikzpicture}
 		[
 		declare function={unipdf(\x,\xl)= (\x>\xl)?1:0.005*rand);}
 		]
 		\begin{axis}[
 		minor tick num=3,
 		axis y line=center,
 		axis x line=middle,
 		enlargelimits = true,
 		xmin = 0,xmax=7,
 		ymin=-50,ymax =30000,
 		ylabel={$N$, epochs}, xlabel={f, Hz},
 		no markers,
 		y tick label style={/pgf/number format/.cd,%
 			scaled y ticks = false,
 			set thousands separator={},
 			fixed},
 		x tick label style={/pgf/number format/.cd,%
 			scaled x ticks = false,
 			set decimal separator={,},
 			fixed}    ]   
 		\addplot [domain=0:6, very thick, blue, samples=100,smooth]{25000*unipdf(x,4.2)*(1+0.005*rand)  };
 		 
 		\end{axis}   
 		
 		\end{tikzpicture}
 		\caption{The relationship between the number of iterations and the frequency of the disturbance. The cutoff frequncy of gyro stabilizer  4 Hz.}
 	\end{center}
 	\label{pic:Epochs}
 \end{figure}
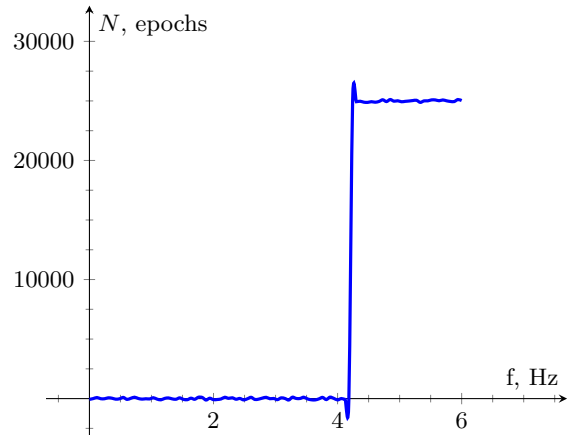

  \begin{figure}[h!]
  	\begin{center}
  		\begin{tikzpicture}
  		[
  		declare function={unipdf(\x,\xl)= (\x>\xl)?1:0;}
  		]
  		\begin{axis}[
  		minor tick num=3,
  		axis y line=center,
  		axis x line=middle,
  		enlargelimits = true,
  		xmin = 0,xmax=7,
  		ymin=-15,ymax =90,
  		ylabel={$N$, epochs}, xlabel={f, Hz},
  		no markers,
  		y tick label style={/pgf/number format/.cd,%
  			scaled y ticks = false,
  			set thousands separator={},
  			fixed},
  		x tick label style={/pgf/number format/.cd,%
  			scaled x ticks = false,
  			set decimal separator={,},
  			fixed}    ]   
  		\addplot [domain=0:6,smooth,very thick, blue] table {
  			0.5 79
  			0.6 75
  			0.8 68
  			0.9 60	
  			1.4 38
  			2.4 24
  			3.4 20
  			4.2 14
  			6.1 9
  		};
  		
  		\end{axis}   
  		\end{tikzpicture}
  		\caption{The relationship between the the maximum angle of the pumping and the frequency of the disturbance. The cutoff frequncy of  gyro stabilizer  4 Hz.}
  	\end{center}
  	\label{pic:freq}
  \end{figure}
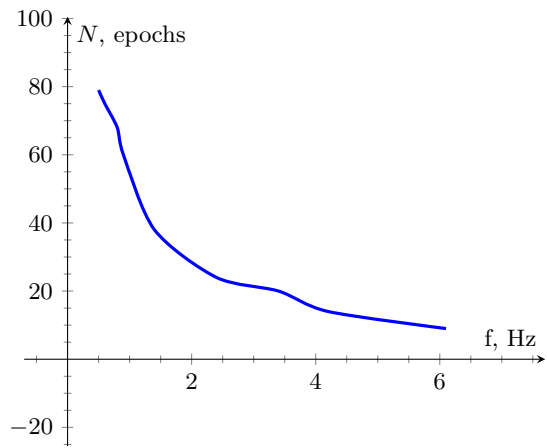

 For determining the optimum frequency of the input harmonic signal numeric experiment was conducted.The input to the reference model supplied harmonic signal with a fixed frequency, after which the obtained sample was trained the neural network. The number of epochs required to train the neural network, and maximum angle leveling platforms comprising a feedback loop neural network was measured in the experiments.

 The results of numeric experiments are shown in Figures \ref{pic:freq} and \label{pic:Epochs}. The figure shows that increasing the frequency of the input harmonic signal decreases the maximum angle pumping platform, but after a certain frequency is a sharp increase in the number of periods required for training. The frequency, then a sharp increasing the number of periods, was close to the cutoff frequency  of the reference model. I Thus, the optimum in terms of the ratio of the time of training and the maximum angle of pumping, is situated at the cutoff frequency.

\subsection{Selecting learning algorithm }

The goal of training the neural network is changing the weight coefficients, which the minimization of  functional \cite{IEEEhowto:Haykin}:

\begin{equation}
E(w)=\frac{1}{2\,N}\Sigma \left(x(t)-\widehat{x}(t,\textbf{w}) \right)^2=
   \frac{1}{2\,N}\Sigma\varepsilon^2,
\label{eq:crit}   
\end{equation}

where $x(t)$ - the training sample,  $\widehat{x}(t,\textbf{w})$ - the output of  neural network,  $N$- the number of training samples.

\begin{itemize}
\item Gradient method;
\item Hewton method;
\item Levenberg-Marquardt method \cite{IEEEhowto:Lera},\cite{IEEEhowto:Ngia}.
\end{itemize}

The learning of neural network consist of several steps. At the begin training set is shuffle. After that  minimization of \label{eq:crit} is doing. These two steps repeat until \label{eq:crit} achieving preset value, or number of loop iteration not be a huge.

As shown by mathematical modeling, the most efficient is the Levenberg-Marquardt algorithm.

\begin{figure}[ht!]
	\begin{center}
		\begin{tikzpicture}
		\begin{axis}[
		minor tick num=3,
		axis y line=center,
		axis x line=middle,
		ymode=log,
		xmode = log,
		enlargelimits = true,
		ymax = 1e5,    
		no markers,
		y tick label style={/pgf/number format/.cd,%
			scaled y ticks = false,
			set thousands separator={},
			fixed},
		x tick label style={/pgf/number format/.cd,%
			scaled x ticks = false,
			precision = 1, 
			set decimal separator={.},
			fixed}
		]   
		\addplot[color=red,samples=500,mark=*,very thick] table[y=y1,x=x] {learn.dat};
		\addlegendentry{Levenrg-Marquardt alorithm};
		\addplot[color=blue,samples=500,mark=*,very thick] table[y=y2,x=x] {learn.dat};
		\addlegendentry{Gradient algorithm};
		\addplot[color=black,samples=500,mark=*,very thick] table[y=y3,x=x] {learn.dat};
		\addlegendentry{Newton algorithm};
		\end{axis}
		\end{tikzpicture}
		\caption{The learning rate}
	\end{center}
	\label{pic:Learn}
\end{figure}
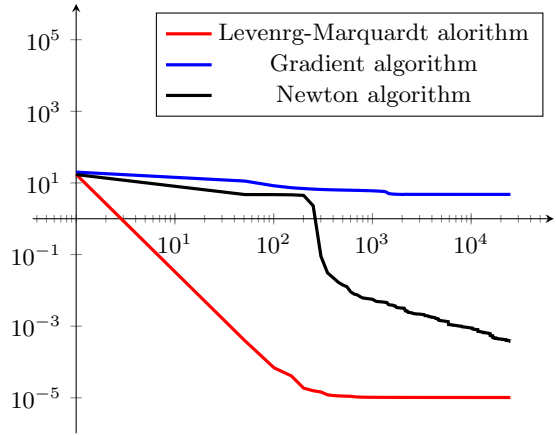

\begin{figure}[h!]
	\centering
	\begin{tikzpicture}
	\plotffind{0}{{t, sec.}}{1}{{$\alpha_1$, arcmin}}{alpha_tansig_1_1_2.dat}
	\end{tikzpicture}
	\caption{Angle plumping when NN(tansig, 1,2)}
\end{figure}
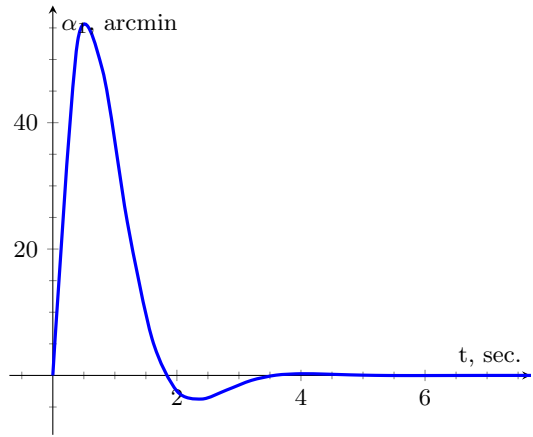

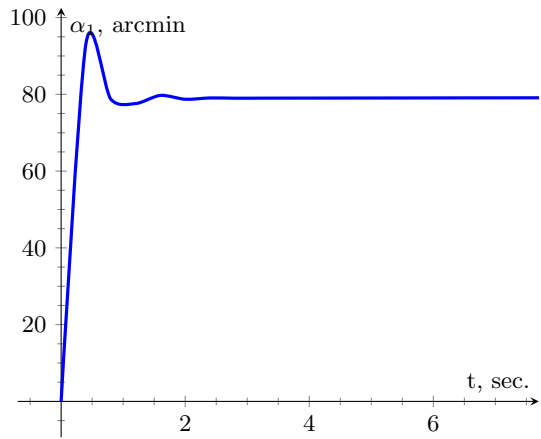
\begin{figure}[h!]
	\centering
	\begin{tikzpicture}
	\plotffind{0}{{t, sec.}}{1}{{$\alpha_1$, arcmin}}{alpha_logsig_1_3_7.dat}
	\end{tikzpicture}
	\caption{Angle plumping when NN(logsig, 3,7)}
\end{figure}

\subsection{Verification of a neural network }

The neural network operates in a feedback loop of a dynamic system, so that traditional methods verification of the neural network are not applicable. In this regard, the only method of verification is a simulation of a closed system. In this case modeling gyro stabilizer  However, in some cases, simulations can take time comparable to the time of training, and even exceed it. In addition to mathematical modeling, it is proposed to check the performance of Neural network at the  stand.

\section{Conclusion}
The article was considered a neural network algorithm for controlling a uniaxial gyro stabilizer. The optimal parameters of neural network based observer are determing. The optimum frequency of the harmonic signal input od ideal model for the formation of a training sample. The results can be used in the synthesis of control devices built using the device of neural networks for tracking systems.

\end{document}